\documentclass[conference]{IEEEtran}
\IEEEoverridecommandlockouts

\usepackage{cite}
\usepackage{amsmath,amssymb,amsfonts}
\usepackage{algorithmic}
\usepackage{graphicx}
\usepackage{textcomp}
\usepackage{booktabs}
\usepackage{xcolor}
\usepackage{multirow}
\usepackage{hyperref}
\def\BibTeX{{\rm B\kern-.05em{\sc i\kern-.025em b}\kern-.08em
    T\kern-.1667em\lower.7ex\hbox{E}\kern-.125emX}}
\begin{document}

\title{TSAI-MetaFraud: A Benchmark Dataset for Financial Fraud Transaction and Behavioral Risk Detection in Metaverse Ecosystems


\thanks{Dataset available at: \url{https://github.com/tsai-unb/MetaFraud}.}
}

\author{\IEEEauthorblockN{Refat Ishrak Hemel, Ehsan Hallaji, and Roozbeh Razavi-Far}
\IEEEauthorblockA{\textit{Trustworthy and Secure AI Lab (TSAI Lab), Faculty of Computer Science, University of New Brunswick, Canada} \\
\{refatishrak.hemel, e.hallaji, roozbeh.razavi-far\}@unb.ca}
}

\maketitle

\begin{abstract}
The emergence of metaverse platforms has created virtual economies that introduce new challenges related to fraud, bot activity, and illicit financial behavior. Despite growing interest in trustworthy metaverse analytics, existing datasets typically focus on user behavior, authentication, or financial transactions in isolation, limiting the development and reproducible evaluation of multimodal fraud detection methods. To address this gap, we present TSAI-MetaFraud, a multimodal, multi-task benchmark dataset for fraud analytics in virtual economies. TSAI-MetaFraud integrates behavioral, transactional, and graph-structured information while incorporating realistic fraud and automated bot scenarios. We define benchmark tasks including transaction fraud detection, cross-modal node classification, temporal link prediction, and weakly supervised fraud detection, and provide baseline evaluations using machine learning models and graph neural networks. By jointly capturing behavioral activity, financial interactions, and relational structure within a unified virtual economy, TSAI-MetaFraud provides a benchmark for advancing multimodal learning, graph mining, fraud analytics, and trustworthy AI in emerging metaverse ecosystems.

\end{abstract}


\section{Introduction}

The rapid development of metaverse technologies is transforming virtual environments into persistent digital ecosystems where users socialize, collaborate, trade virtual assets, and participate in increasingly sophisticated economic activities. Platforms supporting virtual commerce, digital ownership, and avatar-based interactions are expected to play a significant role in future online economies \cite{10.1145/3626315}. As these environments evolve, concerns regarding trust, security, and financial integrity have become increasingly important \cite{10026513, 10550925}. Similar to traditional financial systems, metaverse economies are vulnerable to fraudulent transactions, automated bot activity, account manipulation, identity abuse, and coordinated malicious behavior \cite{10550925,Nair_2023}.

Recent advances in machine learning, graph mining, and fraud analytics have demonstrated considerable success in detecting illicit activities in domains such as banking, e-commerce, and cryptocurrency networks \cite{MOTIE2024122156, Cheng_2025}. However, applying these techniques to metaverse environments remains challenging due to the scarcity of publicly available datasets. Existing fraud datasets primarily focus on transaction networks, while metaverse datasets often emphasize user interactions, virtual environments, or visual content. Consequently, researchers lack realistic benchmarks that jointly capture user behavior, financial activity, and interaction networks within a unified virtual-world setting.

The metaverse introduces unique characteristics that further distinguish it from conventional fraud detection domains. Financial transactions are closely intertwined with avatar behavior, social interactions, and movement patterns \cite{10550925}. Malicious actors may exploit both behavioral and financial channels simultaneously, making it necessary to analyze heterogeneous information sources rather than relying on transactions alone \cite{KARAPATAKIS2025100118}. Furthermore, the dynamic and evolving nature of virtual environments creates complex network structures and temporal dependencies that are not adequately represented in existing benchmarks.

To address these limitations, we introduce TSAI-MetaFraud, a multimodal benchmark dataset collected from a comprehensive metaverse environment built using OpenSimulator. TSAI-MetaFraud captures both behavioral and financial aspects of virtual-world activity through a combination of avatar interactions, transaction records, biometric attributes, and graph-based relationships. The dataset incorporates realistic benign and malicious behaviors, including automated bots, fraudulent transactions, and hybrid attack scenarios, providing a comprehensive testbed for fraud analytics and trustworthy AI research. Beyond releasing the dataset, we establish a suite of benchmark tasks designed to facilitate reproducible evaluation across multiple research communities. These tasks include heterogeneous fraud detection, cross-modal node classification, and temporal link prediction. We further provide baseline results using representative machine learning and Graph Neural Network (GNN) approaches to characterize the challenges posed by the dataset and establish reference performance levels. Hence, the primary contributions of this work are as follows:
\begin{itemize}
\item We introduce TSAI-MetaFraud, a multimodal benchmark that jointly captures behavioral activity, financial transactions, graph-structured interactions, and fraud-related behaviors within a metaverse environment.
\item We define a suite of benchmark tasks spanning fraud detection, bot detection, temporal link prediction, and weakly supervised learning, enabling systematic evaluation of multimodal and graph-based methods.
\item We provide baseline results using machine learning and graph neural network models and publicly release the dataset and evaluation framework to support reproducible research.
\end{itemize}


\section{Related Work}
\label{sec:background}

To position TSAI-MetaFraud within the current research landscape, this section reviews existing metaverse datasets, discusses related benchmarks from neighboring domains, and highlights the limitations of current resources that motivate the proposed benchmark.

\begin{table}[t]
\centering
\caption{Comparison of TSAI-MetaFraud with existing metaverse datasets and related benchmarks from neighboring domains.}
\label{tab:dataset_comparison}
\begin{tabular}{lccccc}
\toprule
Dataset &
\rotatebox{90}{Behavior} &
\rotatebox{90}{Transactions} &
\rotatebox{90}{Fraud Label} &
\rotatebox{90}{Bot Label} &
\rotatebox{90}{Metaverse} \\
\midrule

256-MetaverseRecords \cite{10.1145/3664647.3681711} &
\checkmark & -- & -- & -- & \checkmark \\

MetaPed \cite{metaped} &
\checkmark & -- & -- & -- & \checkmark \\

MGAD \cite{rvh5-8842-25} &
\checkmark & -- & -- & -- & \checkmark \\

Deep Motion Masking \cite{10536254} &
\checkmark & -- & -- & -- & \checkmark \\

Metaverse Financial Transactions \cite{metafinance} &
-- & \checkmark & \checkmark & -- & \checkmark \\

Cresci-15 \cite{cresci2015fame} &
\checkmark & -- & -- & \checkmark & -- \\

Cresci-17 \cite{7876716} &
\checkmark & -- & -- & \checkmark & -- \\

TwiBot-22 \cite{10.5555/3600270.3602825} &
\checkmark & -- & -- & \checkmark & -- \\

Game Bot Detection \cite{kang2016multimodal} &
\checkmark & -- & -- & \checkmark & -- \\

Game-Contagion \cite{8355800} &
\checkmark & -- & -- & \checkmark & -- \\

\textbf{TSAI-MetaFraud (Ours)} &
\checkmark &
\checkmark &
\checkmark &
\checkmark &
\checkmark \\
\bottomrule
\end{tabular}
\vspace{-0.1cm}
\end{table}

\subsection{Metaverse and Virtual-World Datasets}

The growing interest in metaverse technologies has stimulated the creation of datasets spanning virtual economies, user behavior, immersive interactions, security, and privacy. However, existing resources typically focus on a single aspect of virtual environments and do not provide a unified benchmark for studying financial fraud, behavioral anomalies, and relational interactions simultaneously.

One line of research focuses on virtual-world content and behavioral data. The 256-MetaverseRecords dataset \cite{10.1145/3664647.3681711} provides annotated recordings collected from multiple metaverse environments and supports tasks such as event detection, interaction recognition, and scene understanding. Similarly, MetaPed \cite{metaped} employs metaverse simulations to generate synthetic pedestrian behaviors for autonomous driving applications. While these datasets demonstrate the potential of virtual environments for large-scale data generation, they primarily target computer vision tasks and do not contain transactional or security-related information.

A second group of datasets investigates authentication, privacy, and user profiling in immersive environments. The Metaverse Gait Authentication Dataset \cite{rvh5-8842-25} focuses on gait-based biometric authentication using synthetic avatar movements.  Deep Motion Masking \cite{10536254} introduced a dataset for studying privacy-preserving anonymization of VR motion telemetry, while Nair et al. released datasets demonstrating large-scale user identification \cite{291259} and personal attribute inference \cite{10536245} from head and hand motion data. MetaData \cite{Nair_2023} further showed that adversarially designed virtual environments can infer sensitive user characteristics through behavioral responses during gameplay. These datasets have significantly advanced research on privacy, authentication, and behavioral analytics in virtual environments. However, they primarily focus on motion telemetry and user profiling, and do not contain financial transactions, fraud labels, or interaction networks required for studying security threats in virtual economies.

Financially oriented metaverse datasets remain relatively scarce. The Metaverse Financial Transaction Dataset \cite{metafinance} provides transaction records designed for anomaly detection and fraud analysis in virtual economies. Nevertheless, it lacks peer-reviewed benchmark tasks and does not incorporate user behavior, avatar interactions, or relational information beyond transactions.

\subsection{Related Domains}

Although metaverse-specific datasets remain limited, several neighboring domains have contributed benchmarks that capture aspects of user behavior, bot activity, and fraud detection. In social networks, benchmark datasets such as Cresci-15 \cite{cresci2015fame} and Cresci-17 \cite{7876716} have played a central role in advancing social bot detection by providing labeled human and automated accounts exhibiting varying levels of sophistication. More recently, TwiBot-22 \cite{10.5555/3600270.3602825} introduced a large-scale graph-based benchmark that incorporates diverse user attributes, relationships, and interaction structures, facilitating the development of GNN approaches for bot detection.

Related efforts have also emerged in online gaming environments. The Game Bot Detection Dataset \cite{kang2016multimodal} provides long-term behavioral logs collected from a large-scale MMORPG and includes both legitimate players and confirmed bots. Similarly, Game-Contagion \cite{8355800} captures the diffusion of cheating behavior through social interaction networks, enabling the study of behavioral influence and fraud propagation in virtual communities. These datasets demonstrate the importance of combining behavioral signals with relational information when analyzing malicious activity in large-scale digital ecosystems.

Beyond fraud analytics, behavioral datasets have been widely studied in the human activity recognition community \cite{woo2026human}. These benchmarks typically rely on wearable or mobile sensors to classify user activities and have contributed substantially to behavior modeling and representation learning. However, unlike metaverse environments, they do not incorporate virtual interactions, economic transactions, or adversarial behaviors.

Collectively, these neighboring domains provide valuable insights into behavioral analytics, bot detection, and graph-based learning. Nevertheless, none simultaneously capture avatar behavior, financial transactions, interaction networks, fraud labels, and bot activity within a metaverse economy. TSAI-MetaFraud addresses this gap by unifying these components into a single multimodal benchmark specifically designed for fraud analytics and trustworthy AI in virtual environments.

\subsection{Research Gap}

\begin{figure*}[t]
	\centering
    \includegraphics[width=\textwidth]{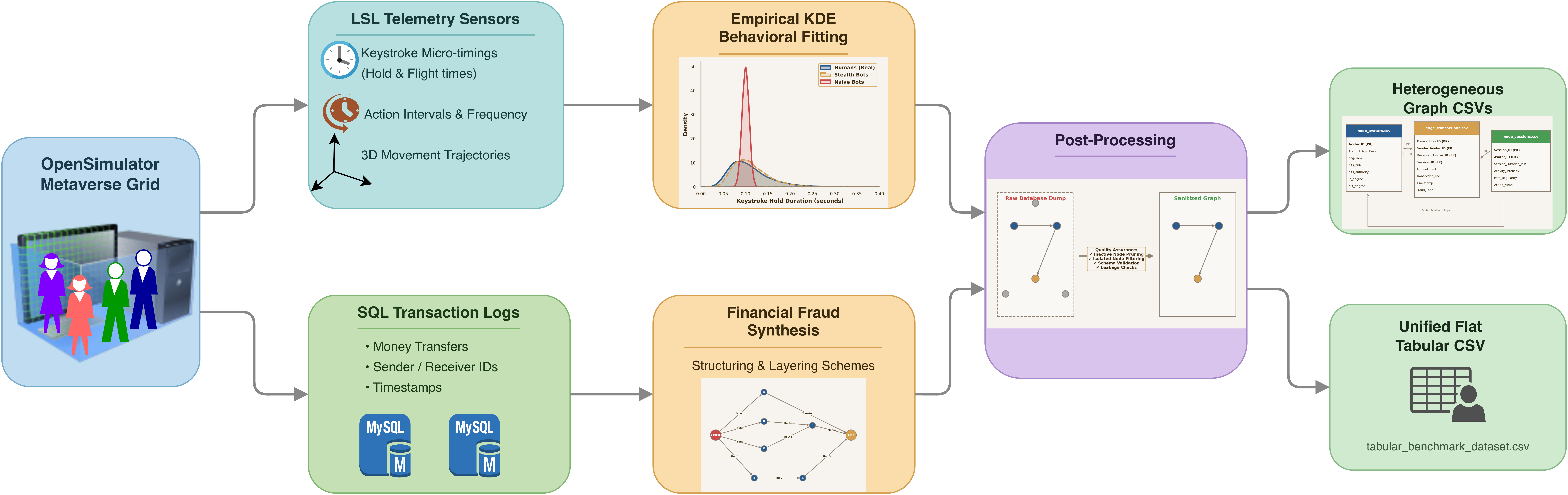}
	\caption{Architecture of the TSAI-MetaFraud data generation pipeline. Behavioral telemetry and financial transactions are collected from the OpenSimulator environment, processed through behavioral modeling and fraud synthesis modules, and transformed into graph and tabular benchmark representations.}
	\label{fig:data_pipeline}
\end{figure*}

Table \ref{tab:dataset_comparison} summarizes the characteristics of existing metaverse, social-network, and online-gaming datasets relevant to fraud analytics and behavioral modeling. While prior datasets provide valuable resources for studying individual aspects of virtual environments, such as user behavior, authentication, bot detection, or financial transactions, none simultaneously combine behavioral activity, financial transactions, fraud annotations, and bot labels within a metaverse setting. This limitation restricts the development and reproducible evaluation of multimodal methods that jointly leverage behavioral and transactional information for fraud analytics. TSAI-MetaFraud addresses this gap by providing a unified benchmark that integrates these components and supports multiple learning tasks, including fraud detection, bot detection, temporal link prediction, and weakly supervised inference within a realistic virtual economy.

The use of a simulated environment is motivated by the limited availability of real-world metaverse fraud datasets, which are often restricted by privacy concerns, platform policies, and the scarcity of verified fraud labels. Simulation enables controlled generation of both benign and malicious behaviors while providing complete ground-truth annotations for reproducible evaluation. Similar simulation-based approaches have been widely adopted in virtual-world, cybersecurity, and autonomous-system research when real-world data are inaccessible or difficult to label.

\section{Dataset Generation and Collection}
\label{sec:dataset_collection}

This section describes the methodology used to construct TSAI-MetaFraud, including the virtual-world environment, data generation pipeline, participant modeling strategy, and post-processing procedures used to produce the final benchmark.

\subsection{Metaverse Environment}
\label{subsec:metaverse_env}

TSAI-MetaFraud was generated within a virtual-world environment built using OpenSimulator (OpenSim), an open-source platform compatible with the Second Life protocol. OpenSim provides a persistent multi-user environment supporting avatar interactions, virtual asset ownership, in-world currency transfers, and user authentication, making it suitable for simulating realistic virtual economies.

The virtual world was designed as a multi-region environment consisting of commercial, trading, and residential areas. Avatars could navigate the environment, interact with virtual objects, purchase items, and perform peer-to-peer financial transactions using the native currency system. These activities generated both behavioral and transactional traces that were recorded throughout the simulation.

The simulated population consisted of 936 active avatars representing different behavioral and financial profiles. Benign users performed normal movements and legitimate financial transactions. Behavioral fraud accounts represented automated bots exhibiting synthetic interaction patterns and abnormal keystroke characteristics. Financial fraud accounts followed realistic movement behaviors while participating in illicit financial activities such as transaction layering and structuring. Hybrid fraud accounts combined automated behavioral patterns with fraudulent financial activity. In addition, a subset of accounts was assigned to an \textit{unknown} category, where labels were intentionally masked to emulate the uncertainty commonly encountered in real-world investigations. The final dataset contains 45 behavioral fraud accounts, 16 financial fraud accounts, 10 hybrid fraud accounts, and 400 unknown accounts, with the remaining avatars representing benign users. 


\subsection{Data Generation Pipeline}
\label{subsec:data_pipeline}

Behavioral and transactional data were collected through a multi-layer acquisition pipeline integrated directly into the OpenSimulator environment, as shown in Fig. \ref{fig:data_pipeline}. Avatar activities were monitored using in-world Linden Scripting Language scripts that recorded movement trajectories, object interactions, and user actions in real time. Simultaneously, server-side transaction logs captured all financial events, including timestamps, sender and receiver identifiers, transaction amounts, transaction types, and associated fees.

To model realistic user behavior, avatar movement events were linked with keystroke timing characteristics. Human behavioral distributions were derived from the public KMT keystroke dynamics dataset, from which hold-time and flight-time statistics were extracted. Gaussian Kernel Density Estimators (KDEs) were then fitted to these timing distributions and used to generate realistic behavioral traces for benign avatars. In contrast, automated accounts were generated using specialized behavioral models designed to emulate varying levels of sophistication.

Two categories of automated accounts were introduced. Naive bots followed rigid timing distributions exhibiting highly predictable interaction patterns, whereas stealth bots sampled from the human-derived KDE distributions while maintaining substantially reduced behavioral variance. This design creates meaningful overlap between benign and malicious behavior and prevents trivial separation of classes. Consequently, successful detection requires models to learn subtle behavioral characteristics rather than relying on simple timing thresholds.

Server-side APIs connected the OpenSimulator database to the data processing framework, enabling automated extraction and synchronization of behavioral events, biometric characteristics, and transaction records.

\subsection{Ethical Considerations}
\label{subsec:ethics}

TSAI-MetaFraud is generated within a fully simulated environment and does not contain personally identifiable information, financial records, or sensitive user data. The behavioral timing distributions used to model human activity were derived from the publicly available KMT keystroke dynamics dataset, which was originally collected under established consent procedures. All financial activities within TSAI-MetaFraud utilize virtual currency and do not correspond to real-world monetary transactions.

\subsection{Data Processing and Quality Assurance}
\label{subsec:data_cleaning}

Following data generation, several preprocessing and quality-control procedures were applied. Inactive accounts that did not participate in behavioral interactions or financial transactions were removed to eliminate isolated nodes from the final graph representation. Missing behavioral information associated with transaction events was represented using zero-padded session-level features in the tabular representation and handled natively within graph-based learning frameworks.

To support temporal analysis, the continuous data stream was partitioned into a sequence of uniform, non-overlapping chronological intervals. This enables capturing a complete activity cycle, while the resulting temporal segmentation enables dynamic graph construction and temporal prediction tasks.

Finally, all features were validated to ensure consistency and realism. Timing variables were constrained to physically meaningful ranges, transaction records were verified for integrity, and graph connectivity was examined to prevent disconnected components and information leakage between temporal partitions.

\section{Dataset Description}
\label{sec:dataset_description}

This section presents the structure, statistical properties, and key characteristics of the TSAI-MetaFraud benchmark. The dataset is designed to support research on fraud analytics, behavioral modeling, and graph-based learning within virtual economies.

\subsection{Data Representation}
\label{subsec:data_schema}

To serve diverse research communities, TSAI-MetaFraud is distributed in a \textit{dual-format} representation: a modular relational graph format for GNNs, and a unified flat tabular format for standard Machine Learning models. 

\subsubsection{Graph Dataset}
The modular graph format represents avatars and sessions as entities, and transactions as directed relationships. This schema comprises four tables summarized in Table~\ref{tab:dataset_schema}: avatar attributes (\texttt{node\_avatars.csv}), session biometrics (\texttt{node\_sessions.csv}), transaction edges (\texttt{edge\_transactions.csv}), and avatar labels (\texttt{node\_avatars\_classes.csv}).

\subsubsection{Tabular Dataset}
To support conventional machine learning models without requiring graph-based processing, we additionally provide a unified tabular representation of the dataset. 
For every transaction edge, we construct a 24-dimensional feature vector by merging the following features.
\begin{enumerate}
    \item \textit{Transaction Features:} Attributes including transaction value, fees, timestamp, type, and the target \texttt{Fraud\_Label} (8 features).
    \item \textit{Sender Avatar Features:} Network centralities and metrics of the sending account (6 features).
    \item \textit{Receiver Avatar Features:} Network centralities and metrics of the receiving account (6 features).
    \item \textit{Sender Session Biometrics:} Keystroke and movement biometrics of the active login session under which the sender executed the payment (4 features).
\end{enumerate}

\begin{table*}[t]
    \centering
    \caption{Attribute schema for avatars, sessions, and transactions in the benchmark dataset}
    \label{tab:dataset_schema}
    \begin{tabular}{clll}
        \toprule
        Description & Field Name & Data Type & Attribute Description \\
        \midrule
        \multirow{7}{*}{Avatar Nodes} &\texttt{Avatar\_ID} & String & Unique UUID identifying the avatar account. \\
        &\texttt{Account\_Age\_Days} & Float & Lifetime of the avatar account (days). \\
        &\texttt{pagerank} & Float & PageRank centrality representing overall transaction activity. \\
        &\texttt{hits\_hub} & Float & HITS Hub score capturing outgoing transaction volume. \\
        &\texttt{hits\_authority} & Float & HITS Authority score capturing incoming transaction volume. \\
        &\texttt{in\_degree} & Float & Normalized in-degree centrality. \\
        &\texttt{out\_degree} & Float & Normalized out-degree centrality. \\
        \midrule
        \multirow{6}{*}{Session Nodes} &\texttt{Session\_ID} & String & Unique identifier for the chronological session window. \\
        &\texttt{Avatar\_ID} & String & The avatar performing the login session. \\
        &\texttt{Session\_Duration\_Min} & Float & Life span of the session (minutes). \\
        &\texttt{Activity\_Intensity} & Float & Number of micro-actions executed per minute. \\
        &\texttt{Path\_Regularity} & Float & Inverse variance of action intervals (path path regularity). \\
        &\texttt{Action\_Mean} & Float & Average action hold duration based on human KDE (seconds). \\
        \midrule
         \multirow{10}{*}{Transaction Edges}&\texttt{Transaction\_ID} & String & Unique transaction edge transaction. \\
        &\texttt{Sender\_Avatar\_ID} & String & ID of the sending account. \\
        &\texttt{Receiver\_Avatar\_ID} & String & ID of the receiving account. \\
        &\texttt{Timestamp} & Float & Raw epoch timestamp. \\
        &\texttt{Amount\_Sent} & Float & Financial value of the transaction. \\
        &\texttt{Transaction\_Type} & String & Category of transaction (e.g. transfer, payment). \\
        &\texttt{Transaction\_Fee} & Float & Fees paid. \\
        &\texttt{Multi\_Wallet\_Indicator} & Integer & Binary indicator (0/1) for multi-wallet aggregation. \\
        &\texttt{Fraud\_Label} & String & Ground truth transaction label. \\
        &\texttt{Session\_ID} & String & Active session under which the payment occurred. \\
        \midrule
        \multirow{2}{*}{Avatar Labels} &\texttt{Avatar\_ID} & String & Unique UUID identifying the avatar account. \\
        &\texttt{Class\_Label} & String & Ground truth avatar label. \\
        \bottomrule
    \end{tabular}
\end{table*}

\subsection{Statistical Overview}
\label{subsec:stat_overview}

Table~\ref{tab:summary} summarizes the main characteristics of the dataset. TSAI-MetaFraud contains 936 active avatars interacting within a simulated metaverse economy during the data collection period. During this interval, the environment generated 74,671 financial transactions and 230,490 behavioral interaction events, producing a rich multimodal benchmark that combines behavioral, transactional, and relational information.

\begin{table}[htbp]
	\centering
	\caption{Summary statistics of the TSAI-MetaFraud dataset}
	\label{tab:summary}
	\begin{tabular}{ll}
		\toprule
		Statistic & Value \\
		\midrule
		Active avatars & 936 \\
		Behavioral sessions & 936 \\
		Behavioral interactions & 230,490 \\
		Financial transactions & 74,671 \\
		Unique transacting nodes & 936 \\
		\bottomrule
	\end{tabular}
\end{table}

\subsection{Distribution Analysis}
\label{subsec:dist_analysis}

\begin{figure}[t]
    \centering
    \includegraphics[width=0.75\columnwidth]{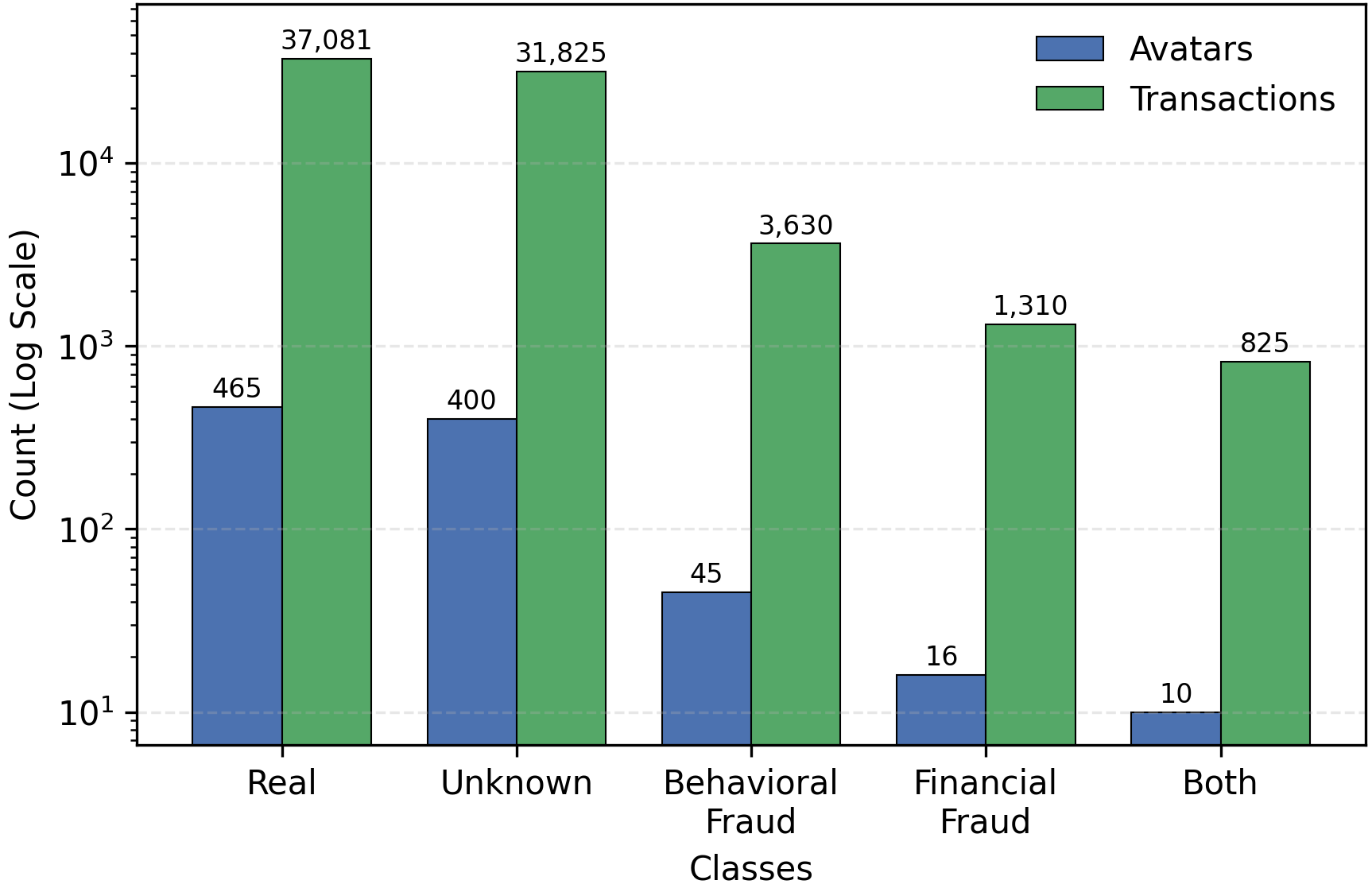}
    \caption{Class distribution of nodes and edge labels, highlighting severe class imbalance.}
    \label{fig:class_dist}
\end{figure}

\begin{figure}[t]
    \centering
    \includegraphics[width=0.77\columnwidth]{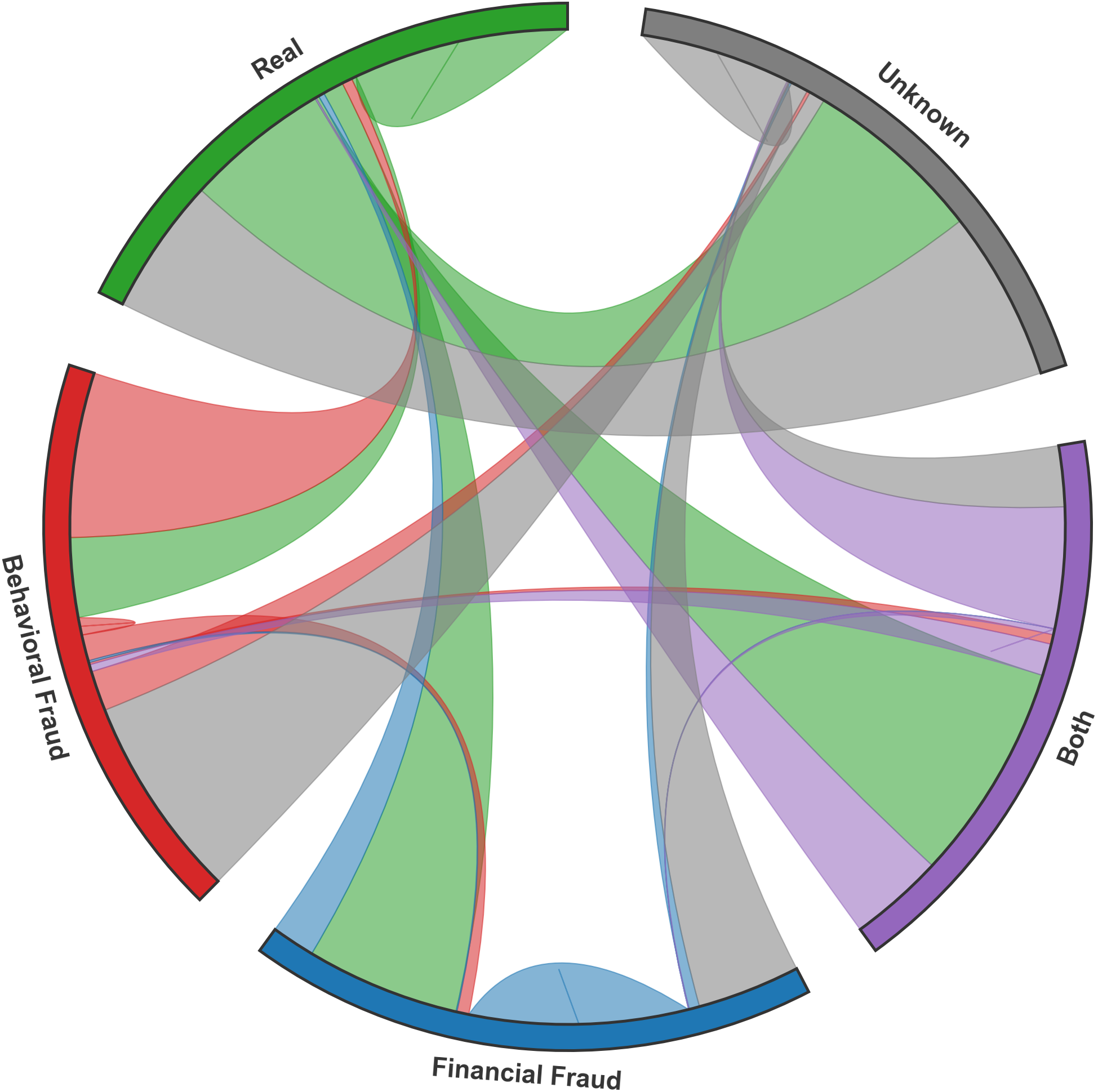}
    \caption{Transaction flow between account categories. Ribbon widths are proportional to the number of transactions exchanged between classes, illustrating interactions among benign, fraudulent, hybrid, and unlabeled entities.}
    \label{fig:chord}
\end{figure}

\begin{figure}[t]
    \centering
    \includegraphics[trim={0.2cm 0.3cm 0.2cm 1.1cm},clip,width=0.85\columnwidth]{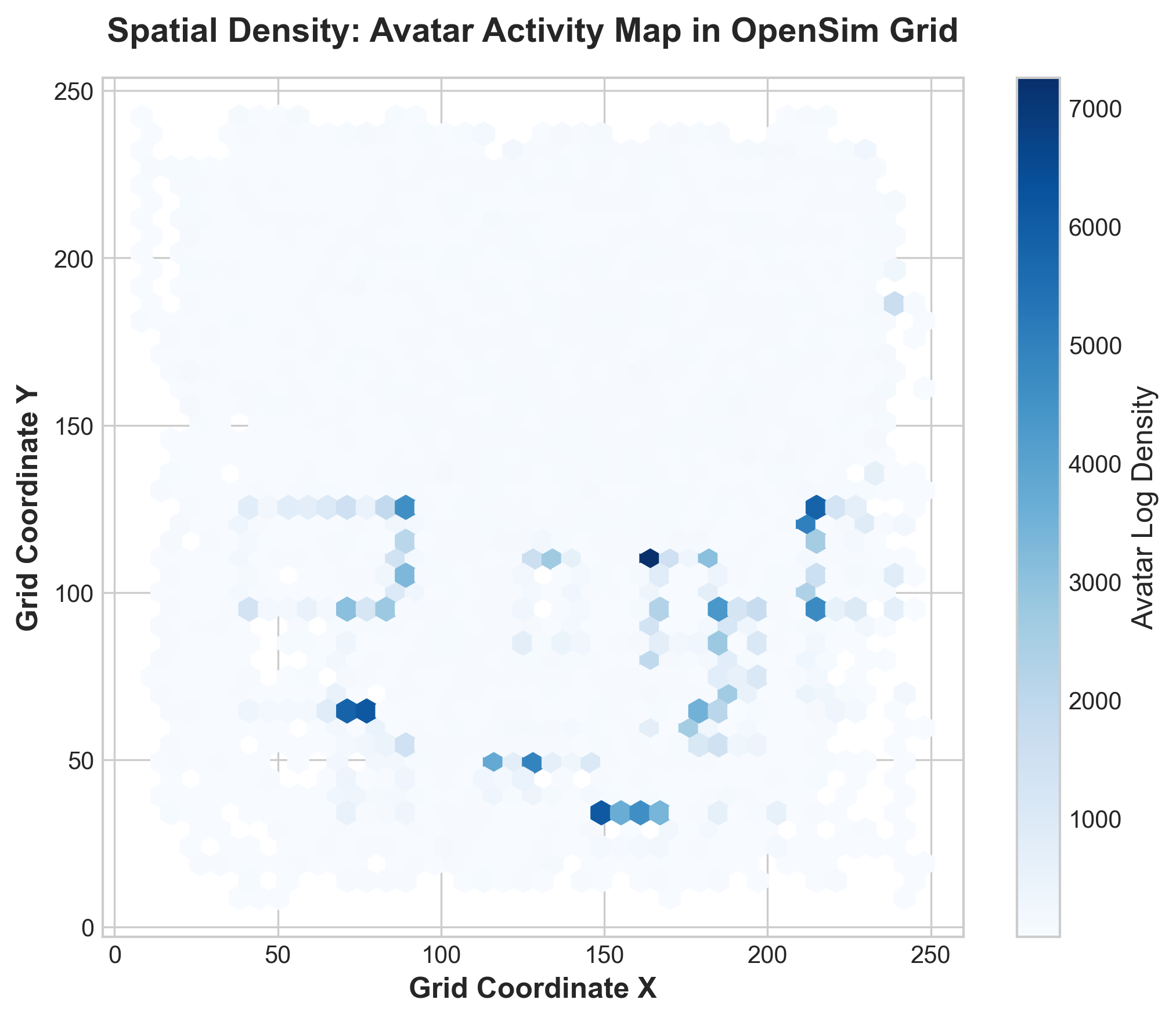}
    \caption{Spatial density heatmap showing avatar coordinates and active hot-spots across the virtual world grid.}
    \label{fig:spatial_dist}
\end{figure}

The dataset exhibits several distributions commonly observed in real-world fraud analytics settings. Fig. \ref{fig:class_dist} presents the class distribution of the benchmark. Similar to many financial crime datasets, TSAI-MetaFraud is highly imbalanced, with benign users representing the majority of the population and fraudulent entities comprising only a small fraction of the network. Fig. \ref{fig:chord} illustrates transaction flows between account categories. While benign and unknown entities account for the majority of transaction activity, fraudulent accounts participate in transactions involving multiple classes rather than forming isolated communities. This cross-class interaction pattern creates a more realistic fraud detection setting in which illicit behavior is embedded within normal economic activity, making detection more challenging than simple community-based identification. 
Fig.~\ref{fig:spatial_dist} visualizes avatar movement patterns within the virtual environment. Activity is concentrated around commercial and trading regions, while peripheral areas remain comparatively sparse. These localized hotspots create realistic spatial heterogeneity and reflect the tendency of economic activity to cluster around specific areas of the virtual world.

\subsection{Network Structure Visualization}
\label{subsec:network_visualization}

\begin{figure*}[t]
    \centering
    \includegraphics[width=\textwidth]{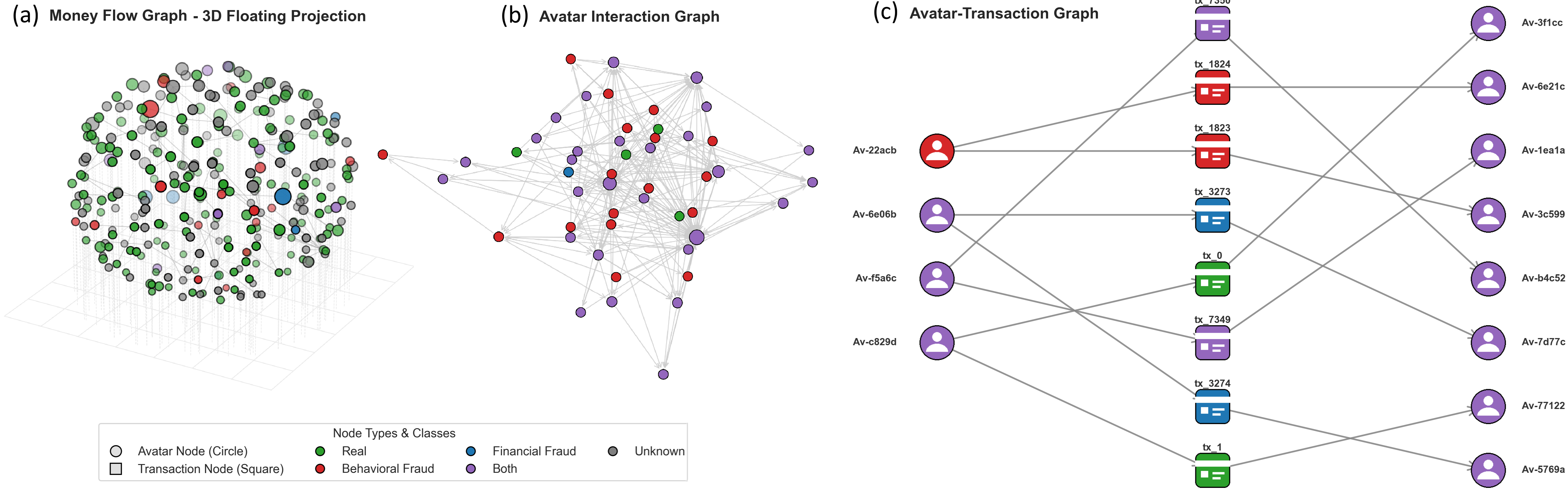}
    \caption{Network representations of TSAI-MetaFraud, including the transaction-to-transaction money flow graph, avatar interaction graph, and avatar--transaction graph. Together, these views illustrate the relational and heterogeneous structures underlying the benchmark.}
    \label{fig:graphs}
\end{figure*}

To illustrate the relational structure of TSAI-MetaFraud, we provide visualizations of the transaction and interaction networks. Fig. ~\ref{fig:graphs}(a) presents the transaction graph connecting sender and receiver accounts through financial transfers. Fig. ~\ref{fig:graphs}(b) illustrates behavioral interactions among avatars, while Fig. ~\ref{fig:graphs}(c) depicts the heterogeneous graph structure integrating avatar entities, behavioral sessions, and transaction relationships. These visualizations highlight the complex topology, sparsity, and heterogeneity of the benchmark and motivate the use of graph-based learning approaches.

\subsection{Data Characteristics}
\label{subsec:data_characteristics}

TSAI-MetaFraud possesses several characteristics that make it a challenging benchmark for machine learning and data mining research.

\subsubsection{Non-Stationarity} Behavioral patterns and transaction activity evolve throughout the simulation period, producing temporal distribution shifts that require models to operate under changing conditions.

\subsubsection{User Heterogeneity} The virtual economy contains users with diverse behavioral profiles, transaction frequencies, and network positions. This results in a heterogeneous graph containing both highly connected hub accounts and sparsely connected participants.

\subsubsection{Graph Sparsity} Despite the large number of transactions, only a small subset of all possible avatar pairs interact directly. Consequently, the resulting transaction network exhibits sparsity characteristics similar to those observed in real financial systems.

\subsubsection{Multimodality} The dataset combines behavioral session information with financial transaction data, enabling the study of methods that jointly leverage behavioral and relational evidence for fraud detection.

\subsubsection{Relational Structure} Financial transactions are naturally represented as directed interactions between entities, allowing the dataset to support graph-based tasks such as node classification, link prediction, and fraud propagation analysis.

\section{Benchmark Tasks}
\label{sec:benchmark_tasks}

To facilitate reproducible evaluation and comparison of future methods, TSAI-MetaFraud defines a suite of benchmark tasks spanning fraud analytics, behavioral modeling, graph learning, and semi-supervised inference. Collectively, these tasks evaluate a model's ability to detect illicit financial activities, identify stealthy automated accounts, forecast future transactions, and operate under realistic conditions of limited supervision. Table~\ref{tab:benchmark_tasks} summarizes the inputs, prediction targets, and evaluation metrics associated with each task.

\begin{table*}[t]
	\centering
	\caption{Benchmark tasks supported by TSAI-MetaFraud.}
	\label{tab:benchmark_tasks}
		\begin{tabular}{c p{3.9cm} p{6.3cm} p{5.5cm}}
			\toprule
			\# & Task &
			Input Data &
			Prediction Target \\
			\midrule
			1 & Transaction fraud detection &
			Transaction attributes and graph-based account features &
			Benign, Behavioral Fraud, Financial Fraud, Both \\
			2& Cross-modal node classification &
			Features of behavioral session and account-level graph &
			Benign, Behavioral Fraud, Financial Fraud, Both  \\
			3 & Temporal link prediction &
			Sequence of temporal transaction graphs &
			Future transaction formation probability \\
			4 & Weakly supervised fraud detection &
			Partially labeled transaction graph &
			Multiclass fraud labels of unlabeled accounts \\
			\bottomrule
		\end{tabular}
	
\end{table*}

\subsection{Transaction Fraud Detection}

This task evaluates multiclass fraud detection under a strict inductive setting. Models are required to classify transaction edges involving previously unseen accounts using both transactional attributes and graph-derived structural information. The objective is to classify each transaction into one of four categories: normal transactions (Benign), Behavioral Fraud, Financial Fraud, or hybrid (Both), while generalizing to new participants that were not observed during training.

\subsection{Cross-Modal Node Classification}

This task focuses on identifying avatar profiles using multimodal behavioral and structural information. Session-level behavioral features (i.e., from login sessions) are combined with graph-derived account characteristics (centralities) to classify avatar nodes into one of the four categories: real humans (Benign), automated scripts (Behavioral Fraud), illicit financial accounts (Financial Fraud), and hybrid automated fraud accounts (Both). By evaluating models on stealthy bot profiles and illicit financial flows concurrently, this task measures a model's ability to generalize across orthogonal data modalities.

\subsection{Temporal Link Prediction}

This task evaluates a model's ability to forecast future transactions in an evolving financial network. Given a sequence of historical transaction graph snapshots, the objective is to predict whether a directed transaction will occur between two accounts in a future time interval. The task captures the temporal and non-stationary nature of activity within virtual economies.

\subsection{Weakly Supervised Fraud Detection}

This task simulates realistic fraud investigation scenarios in which only a small subset of entities is labeled. Models must leverage graph structure and behavioral information to infer the multiclass label (i.e., similar to the four defined labels of previous tasks) of a largely unlabeled population. By masking the majority of available labels (90\% masked), the benchmark evaluates the ability of learning algorithms to operate effectively under severe label scarcity.

\section{Evaluation and Analysis}
\label{sec:results}

To characterize the difficulty of the proposed benchmark and establish reference performance levels for future research, we evaluate a diverse set of baseline models spanning classical machine learning, graph representation learning, and graph-based label propagation. These baselines were selected to reflect common approaches used in fraud analytics, bot detection, and network mining.

\subsection{Baseline Models}

\begin{table}[t]
	\centering
	\caption{Baseline models used for evaluating TSAI-MetaFraud.}
	\label{tab:baseline_models}
	\begin{tabular}{llc}
		\toprule
		Model & Category & Data\\
		\midrule
		  Logistic Regression & Linear Classification & Tabular \\
		  Random Forest & Tree-Based Learning & Tabular\\
		XGBoost & Gradient Boosting & Tabular\\
		GraphSAGE (Node Classifier) & Graph Neural Networks & Graph\\
		GraphSAGE (Link Predictor) & Temporal Graph Learning & Graph\\
		Label Propagation & Graph Diffusion & Graph\\
		\bottomrule
	\end{tabular}
\end{table}

The evaluated baselines are summarized in Table~\ref{tab:baseline_models}. Logistic Regression, Random Forest \cite{breiman2001random}, and XGBoost \cite{10.1145/2939672.2939785} are trained on the tabular representation of the dataset and serve as representative classical baselines. These methods provide insight into the predictive value of handcrafted behavioral and transactional features without explicitly modeling graph structure.

To evaluate graph-based representation learning, we employ a heterogeneous GraphSAGE model operating on the multi-relational transaction network \cite{10.5555/3294771.3294869}. The model propagates information across avatar and session entities through transactional and behavioral relationships, enabling node-level and edge-level prediction tasks. For temporal link prediction, a GraphSAGE encoder coupled with a link prediction head is applied to sequences of transaction graph snapshots. Finally, a standard Label Propagation algorithm is included as a non-parametric graph diffusion baseline for weakly supervised fraud detection.

\subsection{Evaluation Protocol}

To prevent information leakage and better reflect realistic deployment scenarios, task-specific evaluation protocols are employed. For transaction fraud detection and behavioral bot detection, a strict inductive split is adopted in which unique avatars are partitioned into training (80\%) and testing (20\%) sets. Consequently, all test entities remain unseen during training. For temporal link prediction, transaction records are divided into chronological one-hour snapshots. Models are trained on the historical sequence of transaction snapshots preceding the final observation window, and evaluated on their ability to forecast links that emerge in the subsequent, unseen period. Negative examples are generated through random sampling of non-existent transaction pairs. For weakly supervised fraud detection, the inductive split is retained while additionally masking 90\% of the available training labels, simulating realistic conditions in which only a small subset of suspicious accounts can be manually investigated.

\subsection{Experimental Setup}
\label{subsec:experimental_setup}

To prevent information leakage and evaluate realistic cold-start performance, we adopt a strict inductive split based on unique \texttt{Avatar\_ID}s. Specifically, 80\% of avatars are assigned to the training set, while the remaining 20\% are reserved for testing. Transaction edges are partitioned accordingly: the training set contains 34,253 transactions for which both sender and receiver belong to the training avatar set, whereas the inductive test set contains 8,593 transactions involving at least one previously unseen avatar. This protocol ensures that models are evaluated on their ability to generalize to new participants rather than memorizing previously observed accounts. For classical machine learning baselines, hyperparameter selection is performed using five-fold cross-validation on the training split. GNN models are trained for 100 epochs using cross-entropy loss.

All experiments were conducted using Python. Graph-based models were implemented using PyTorch and PyTorch Geometric, while classical baselines were implemented using Scikit-Learn and XGBoost. Experiments were executed on a workstation equipped with an Apple Silicon processor and 16GB of unified memory.

Given the highly imbalanced nature of fraud detection tasks, we report class-specific Precision, Recall, and F1-score, together with overall Macro F1-score. Macro-averaged metrics are particularly important as they provide a balanced evaluation across both majority and minority classes.

\subsection{Main Results}
\label{subsec:main_results}

The objective of the baseline evaluation is not to establish state-of-the-art performance, but rather to characterize the difficulty of the proposed benchmark and provide reference results for future research. The evaluated methods span classical tabular learning, graph representation learning, temporal link prediction, and semi-supervised graph inference.

Table~\ref{tab:baseline_performance_tx} presents the results for transaction fraud detection (Task~1) under the strict inductive setting, where transactions involving previously unseen avatars are reserved for testing. The results reveal a clear distinction between behavioral and financial fraud detection. While Random Forest and XGBoost achieve strong performance on the Behavioral Fraud class, they completely fail to identify Financial Fraud and hybrid fraud transactions. In contrast, the heterogeneous GraphSAGE model achieves meaningful performance across all fraud categories, suggesting that graph structure provides critical information for detecting complex illicit transaction patterns.

\begin{table}[t]
	\centering
	\caption{Comparative baseline performance on the inductive test set for transaction classification (Task 1)}
	\label{tab:baseline_performance_tx}
	\begin{tabular}{llccc}
		\toprule
		Model (Data) & Target Class & Precision & Recall & F1-Score \\
		\midrule
		\multirow{2}{*}{Logistic}& Behavioral Fraud & 0.81 & 0.69 & 0.75 \\
		\multirow{2}{*}{Regression}& Both (Hybrid) & 0.00 & 0.00 & 0.00 \\
		  \multirow{2}{*}{(Tabular)}& Financial Fraud & 0.02 & 0.20 & 0.03 \\
		& Real (Benign) & 0.96 & 0.62 & 0.76 \\
		\midrule
		& Behavioral Fraud & 0.86 & 1.00 & 0.93 \\
		Random Forest & Both (Hybrid) & 0.00 & 0.00 & 0.00 \\
		(Tabular) & Financial Fraud & 0.00 & 0.00 & 0.00 \\
		& Real (Benign) & 0.97 & 1.00 & 0.98 \\
		\midrule
		& Behavioral Fraud & 0.81 & 0.69 & 0.75 \\
		XGBoost & Both (Hybrid) & 0.00 & 0.00 & 0.00 \\
		(Tabular) & Financial Fraud & 0.00 & 0.00 & 0.00 \\
		& Real (Benign) & 0.97 & 0.99 & 0.98 \\
		\midrule
		& Behavioral Fraud & 0.34 & 0.49 & 0.40 \\
		GraphSAGE & Both (Hybrid) & 0.55 & 0.62 & 0.58 \\
		(Graph) & Financial Fraud & 0.37 & 0.41 & 0.39\\
		& Real (Benign) & 0.93 & 0.88 & 0.90 \\
		\bottomrule
	\end{tabular}
\end{table}

Table~\ref{tab:baseline_performance_node} reports the results for cross-modal node classification (Task~2). Tree-based methods achieve the strongest overall performance on benign accounts (Real) and automated scripts (Behavioral Fraud), with Random Forest obtaining F1-scores of 0.98 and 0.90 respectively. This indicates that session-level behavioral features (e.g., average action hold durations, path regularity) provide highly discriminative signals for identifying basic automated anomalies. However, all tabular baselines struggle to identify Financial Fraud and hybrid nodes, as their malicious patterns are expressed through multi-hop transactional relationships which are unavailable to standard models.

\begin{table}[t]
	\centering
	\caption{Comparative baseline performance on the inductive test set for Avatar node classification (Task 2)}
	\label{tab:baseline_performance_node}
    \setlength{\tabcolsep}{4pt}
	\begin{tabular}{lllccc}
		\toprule
		Data & Model & Target Class & Precision & Recall & F1-Score \\
		\midrule
		\multirow{4}{*}{Tabular} & & Real (Benign) & 0.95 & 0.66 & 0.78 \\
		& Logistic & Behavioral Fraud & 0.89 & 0.89 & 0.89 \\
		& Regression& Financial Fraud & 0.00 & 0.00 & 0.00 \\
		& & Both (Hybrid) & 0.50 & 0.50 & 0.50 \\
		\midrule
		\multirow{4}{*}{Tabular} &  & Real (Benign) & 0.97 & 1.00 & 0.98 \\
		& Random & Behavioral Fraud & 0.82 & 1.00 & 0.90 \\
		& Forest & Financial Fraud & 0.00 & 0.00 & 0.00 \\
		& & Both (Hybrid) & 0.00 & 0.00 & 0.00 \\
		\midrule
		\multirow{4}{*}{Tabular} & \multirow{4}{*}{XGBoost} & Real (Benign) & 0.96 & 1.00 & 0.98 \\
		& & Behavioral Fraud & 0.80 & 0.89 & 0.84 \\
		& & Financial Fraud & 0.00 & 0.00 & 0.00 \\
		& & Both (Hybrid) & 0.00 & 0.00 & 0.00 \\
		\midrule
		\multirow{4}{*}{Graph} & \multirow{4}{*}{GraphSAGE} & Real (Benign) & 0.89 & 0.78 & 0.83 \\
		& & Behavioral Fraud & 0.11 & 0.22 & 0.14 \\
		&  & Financial Fraud & 0.00 & 0.00 & 0.00 \\
		& & Both (Hybrid) & 0.00 & 0.00 & 0.00 \\
		\bottomrule
	\end{tabular}
\end{table}

Table~\ref{tab:baseline_performance_link} summarizes the results for temporal link prediction (Task~3). Traditional topological heuristics such as Common Neighbors, Jaccard Coefficient, and Preferential Attachment perform poorly in forecasting future transactions. GraphSAGE substantially outperforms these methods, demonstrating the importance of learned structural representations for modeling dynamic transaction networks.

\begin{table}[t]
	\centering
	\caption{Comparative baseline performance on dynamic temporal link prediction (Task 3)}
	\label{tab:baseline_performance_link}
	\begin{tabular}{clcc}
		\toprule
		Data & Model & AUROC & Average Precision \\
		\midrule
		Graph & Common Neighbors & 0.2358 & 0.4051 \\
		Graph & Jaccard Coefficient & 0.1008 & 0.3288 \\
		Graph & Preferential Attachment & 0.3470 & 0.4611 \\
		Graph & GraphSAGE & 0.6089 & 0.6485 \\
		\bottomrule
	\end{tabular}
\end{table}

Table~\ref{tab:baseline_performance_weak} presents the macro-averaged baseline performance on weakly supervised multiclass fraud propagation (Task~4) under extreme label scarcity, where 90\% of training labels are masked. All methods experience a substantial performance degradation compared to the fully supervised setting. Interestingly, Label Propagation and GraphSAGE show limited effectiveness compared to Logistic Regression under severe label scarcity. This behavior is largely driven by the difficulty of identifying the highly imbalanced and extremely sparse minority classes (e.g., Financial Fraud and Both), for which only a handful of labeled examples are available during training. These results highlight the challenges of fraud detection under realistic low-label conditions and motivate the development of more robust label-efficient graph learning methods for metaverse security.

\begin{table}[t]
	\centering
	\caption{Macro-averaged baseline performance on weakly-supervised multiclass fraud propagation (Task 4)}
	\label{tab:baseline_performance_weak}
	\begin{tabular}{llccc}
		\toprule
		Data & Model & Precision & Recall & F1-Score \\
		\midrule
		Tabular & Logistic Regression & 0.4911 & 0.5071 & 0.4873 \\
		Tabular & Random Forest & 0.4423 & 0.4722 & 0.4566 \\
		Tabular & XGBoost & 0.4398 & 0.4722 & 0.4553 \\
		Graph & Label Propagation & 0.2176 & 0.2500 & 0.2327 \\
		Graph & GraphSAGE & 0.2359 & 0.2405 & 0.2377 \\
		\bottomrule
	\end{tabular}
\end{table}

\subsection{Discussion and Insights}
\label{subsec:difficulty_analysis}

The baseline results reveal several characteristics that make TSAI-MetaFraud a challenging benchmark for fraud analytics and graph learning. First, the results highlight a clear distinction between behavioral and financial fraud detection. Tree-based models such as Random Forest and XGBoost achieve strong performance on classifying benign users (Real) and automated scripts (Behavioral Fraud). This suggests that session-level behavioral features, including activity intensity and timing statistics, provide highly discriminative signals for identifying basic automated behavior. However, these same models fail to detect Financial Fraud and hybrid (Both) node categories, achieving zero F1-scores on both classes. This observation indicates that financial fraud patterns are primarily expressed through transaction structures and multi-hop interaction flows rather than local account attributes.

Additionally, GraphSAGE node classifier struggles on Behavioral Fraud (0.14 F1-score) compared to tree baselines. This highlights an important cross-modal difficulty: while graph neural networks are well-suited for topological propagation, the standard message-passing mechanism tends to aggregate feature representations over the neighborhood, which smooths out the local, high-frequency keystroke timing anomalies that define automated behavior.

The temporal link prediction results demonstrate the limitations of traditional topological heuristics in evolving virtual economies. Common Neighbors, Jaccard Coefficient, and Preferential Attachment achieve poor predictive performance, while GraphSAGE attains substantially higher AUROC and Average Precision scores. These results suggest that future transaction formation depends on complex structural and temporal dependencies that cannot be captured using static proximity measures alone.

Finally, the weakly-supervised setting (Task 4) exposes the combined limits of graph diffusion and label scarcity. Under this condition, the rare classes are represented by only a single anchor node. Standard semi-supervised GNNs and Label Propagation struggle to propagate signals effectively from these single points without introducing massive noise, resulting in macro F1-scores of approximately 0.23--0.24. This highlights the severe challenges posed by the long-tailed class distribution when supervision is sparse, establishing TSAI-MetaFraud as a rigorous testbed for few-shot and weakly-supervised graph learning.

\subsection{Modality Analysis}
\label{subsec:ablation_studies}

TSAI-MetaFraud combines two complementary sources of information: behavioral session characteristics and financial transaction networks. The baseline results provide insight into the relative contribution of each modality. Behavioral features prove highly effective for identifying automated scripts. Models trained on session-level characteristics are able to identify Behavioral Fraud and Real accounts with high accuracy, indicating that behavioral signals contain strong discriminative information. However, these same models struggle to identify structural financial fraud, suggesting that illicit transaction routing patterns are not easily observable from local behavioral features alone. Conversely, graph-based methods are better suited for capturing financial fraud and hybrid fraud activities. By leveraging transaction relationships and network structure, GraphSAGE achieves meaningful performance on fraud classes that are largely missed by tabular baselines. This demonstrates the importance of relational information when modeling illicit financial behavior. Taken together, these results suggest that behavioral and transactional modalities provide complementary information. Behavioral features facilitate bot detection, whereas graph structure is essential for identifying complex money-laundering patterns. Consequently, TSAI-MetaFraud provides a valuable benchmark for evaluating multimodal learning approaches that jointly exploit behavioral and relational information.

\section{Discussion}
\label{sec:discussion}
Beyond the benchmark tasks presented in this work, TSAI-MetaFraud enables several research directions in fraud analytics, graph learning, and trustworthy AI, while also presenting limitations that motivate future extensions of the dataset.

\subsection{Research Opportunities}
TSAI-MetaFraud enables several research directions at the intersection of graph mining, fraud analytics, and virtual-world security.

\subsubsection{Multimodal Fraud Detection}
The benchmark combines behavioral telemetry, transactional records, and relational network information within a unified framework. This creates opportunities for developing multimodal learning methods capable of jointly modeling user behavior and financial activity to identify sophisticated fraud patterns.

\subsubsection{Dynamic and Evolving Virtual Economies}
Unlike many existing fraud datasets that provide static snapshots, TSAI-MetaFraud contains temporally evolving interactions and transaction networks. This supports research on temporal graph learning, online fraud detection, and adaptive learning methods capable of operating under non-stationary environments.

\subsubsection{Learning Under Limited Supervision}
The severe class imbalance and label scarcity exhibited by the benchmark reflect practical challenges encountered in fraud investigation. Consequently, TSAI-MetaFraud provides a testbed for semi-supervised learning, graph-based label propagation, active learning, and label-efficient fraud detection techniques.

\subsubsection{Benchmarking Multimodal Graph Learning}
TSAI-MetaFraud provides a unified benchmark for evaluating models that combine behavioral, transactional, and relational information. Such multimodal graph learning approaches have attracted growing interest in fraud detection, financial analytics, and trustworthy AI, yet suitable benchmark datasets remain limited. TSAI-MetaFraud enables systematic evaluation of feature fusion, representation learning, and GNN architectures under realistic conditions involving class imbalance, temporal evolution, and limited supervision.

\subsection{Limitations}
TSAI-MetaFraud represents a specific virtual-world ecosystem implemented within OpenSimulator and may not capture all behaviors and economic dynamics observed across different metaverse platforms. In addition, the benchmark focuses on behavioral bot activity, transaction anomalies inspired by classic money laundering typologies and cryptocurrency financial crimes, and therefore does not encompass every possible form of malicious behavior. Despite these limitations, TSAI-MetaFraud provides a unique benchmark that combines behavioral telemetry, financial transactions, and graph structure within a unified framework, enabling the evaluation of fraud detection, graph learning, and multimodal analytics methods in emerging virtual economies.

\section{Conclusion}
\label{sec:conclusion}

This paper introduced TSAI-MetaFraud, a multimodal benchmark dataset for fraud analytics in virtual economies. The dataset combines behavioral telemetry, financial transactions, and heterogeneous graph structures within a unified framework and is released in both graph and tabular formats. To support reproducible evaluation, we defined four benchmark tasks spanning transaction fraud detection, cross-modal node classification, temporal link prediction, and weakly supervised fraud detection. This dataset fills an important benchmarking gap by providing a unified platform for evaluating multimodal learning, graph mining, and fraud analytics methods in virtual economies, thereby supporting reproducible research in an emerging and rapidly evolving application domain. Baseline experiments demonstrate that the benchmark presents significant challenges arising from multimodality, graph structure, temporal dynamics, class imbalance, and limited supervision. We hope TSAI-MetaFraud will facilitate future research on graph learning, multimodal analytics, and security in emerging virtual-world ecosystems.

\bibliographystyle{IEEEtran}
\bibliography{main}

@inproceedings{10.1145/3664647.3681711,
	author = {Steinert, Patrick and Wagenpfeil, Stefan and Frommholz, Ingo and Hemmje, Matthias L.},
	title = {256 Metaverse Records Dataset},
	year = {2024},
	isbn = {9798400706868},
	booktitle = {Proceedings of the 32nd ACM International Conference on Multimedia},
	pages = {4256–4263},
	numpages = {8}
}

@article{metaped,
	title={Fully Synthetic Pedestrian Anomaly Behavior Dataset Generation in Metaverse for Enhancing Autonomous Driving Object Detection},
	author={Aung, Nang Htet Htet and Sangwongngam, Paramin and Jintamethasawat, Rungroj and Wuttisittikulkij, Lunchakorn},
	journal={IEEE Access},
	volume={12},
	pages={166630--166642},
	year={2024},
	doi={10.1109/ACCESS.2024.3495505}
}

@data{rvh5-8842-25,
	doi = {10.21227/rvh5-8842},
	url = {https://dx.doi.org/10.21227/rvh5-8842},
	author = {Sandeep Ravikanti},
	publisher = {IEEE Dataport},
	title = {Metaverse Gait Authentication Dataset {(MGAD)}},
	year = {2025} }

@INPROCEEDINGS{10536254,
	author={Nair, Vivek and Guo, Wenbo and O'Brien, James F. and Rosenberg, Louis and Song, Dawn},
	booktitle={2024 IEEE Conference on Virtual Reality and 3D User Interfaces Abstracts and Workshops (VRW)}, 
	title={Deep Motion Masking for Secure, Usable, and Scalable Real-Time Anonymization of Ecological Virtual Reality Motion Data}, 
	year={2024},
	volume={},
	number={},
	pages={493-500}}

@inproceedings {291259,
	author = {Vivek Nair and Wenbo Guo and Justus Mattern and Rui Wang and James F. O{\textquoteright}Brien and Louis Rosenberg and Dawn Song},
	title = {Unique Identification of 50,000+ Virtual Reality Users from Head \& Hand Motion Data},
	booktitle = {32nd USENIX Security Symposium (USENIX Security 23)},
	year = {2023},
	isbn = {978-1-939133-37-3},
	pages = {895--910}
}

@INPROCEEDINGS {10536245,
	author = { Nair, Vivek and Rack, Christian and Guo, Wenbo and Wang, Rui and Li, Shuixian and Huang, Brandon and Cull, Atticus and O'Brien, James F. and Latoschik, Marc and Rosenberg, Louis and Song, Dawn },
	booktitle = {IEEE Conference on Virtual Reality and 3D User Interfaces Abstracts and Workshops (VRW) },
	title = {Inferring Private Personal Attributes of Virtual Reality Users from Ecologically Valid Head and Hand Motion Data},
	year = {2024},
	volume = {},
	ISSN = {},
	pages = {477-484}
	}

@data{metafinance,
	title={Metaverse Financial Transaction Dataset},
	author={Jack Ward},
	year={2023},
	publisher={Kaggle},
	url={https://www.kaggle.com/datasets/faizaniftikharjanjua/metaverse-financial-transactions-dataset},
}

@article{Nair_2023,
	title={Exploring the Privacy Risks of Adversarial {VR} Game Design},
	volume={2023},
	ISSN={2299-0984},
	number={4},
	journal={Proceedings on Privacy Enhancing Technologies},
	author={Nair, Vivek and Munilla Garrido, Gonzalo and Song, Dawn and O’Brien, James},
	year={2023},
	pages={238–256} 
	}

@inproceedings{10.5555/3600270.3602825,
	author = {Feng, Shangbin and Tan, Zhaoxuan and Wan, Herun and Wang, Ningnan and Chen, Zilong and Zhang, Binchi and Zheng, Qinghua and Zhang, Wenqian and Lei, Zhenyu and Yang, Shujie and Feng, Xinshun and Zhang, Qingyue and Wang, Hongrui and Liu, Yuhan and Bai, Yuyang and Wang, Heng and Cai, Zijian and Wang, Yanbo and Zheng, Lijing and Ma, Zihan and Li, Jundong and Luo, Minnan},
	title = {TwiBot-22: towards graph-based twitter bot detection},
	year = {2022},
	isbn = {9781713871088},
	booktitle = {Proceedings of the 36th International Conference on Neural Information Processing Systems},
	articleno = {2555},
	numpages = {16}
}

@article{cresci2015fame,
	title={Fame for sale: Efficient detection of fake Twitter followers},
	author={Cresci, Stefano and Di Pietro, Roberto and Petrocchi, Marinella and Spognardi, Angelo and Tesconi, Maurizio},
	journal={Decision Support Systems},
	volume={80},
	pages={56--71},
	year={2015},
	publisher={Elsevier}
}

@ARTICLE{7876716,
	author={Cresci, Stefano and Pietro, Roberto Di and Petrocchi, Marinella and Spognardi, Angelo and Tesconi, Maurizio},
	journal={IEEE Transactions on Dependable and Secure Computing}, 
	title={Social Fingerprinting: Detection of Spambot Groups Through DNA-Inspired Behavioral Modeling}, 
	year={2018},
	volume={15},
	number={4},
	pages={561-576}
	}

@article{kang2016multimodal,
	title={Multimodal game bot detection using user behavioral characteristics},
	author={Kang, Ah Reum and Jeong, Seong Hoon and Mohaisen, Aziz and Kim, Huy Kang},
	journal={SpringerPlus},
	volume={5},
	number={1},
	pages={523},
	year={2016},
	publisher={Springer}
}

@ARTICLE{8355800,
	author={Woo, Jiyoung and Kang, Sung Wook and Kim, Huy Kang and Park, Juyong},
	journal={IEEE Access}, 
	title={Contagion of Cheating Behaviors in Online Social Networks}, 
	year={2018},
	volume={6},
	number={},
	pages={29098-29108}}

@article{woo2026human,
		title={Human Activity Recognition Dataset for Pedestrians with Mobility Disabilities},
		author={Woo, Yeji and Hwang, Sungjin and Oh, Seungwoo and Kang, Myungwon and Lee, Sungyoon and Kim, Jieun and Cha, Jaehyuk and Kim, Kwanguk Kenny},
		journal={Scientific Data},
		year={2026},
		publisher={Nature Publishing Group UK London}
	}

@inproceedings{10.5555/3294771.3294869,
author = {Hamilton, William L. and Ying, Rex and Leskovec, Jure},
title = {Inductive representation learning on large graphs},
year = {2017},
isbn = {9781510860964},
booktitle = {Proceedings of the 31st International Conference on Neural Information Processing Systems},
pages = {1025–1035},
numpages = {11},
}

@inproceedings{10.1145/2939672.2939785,
author = {Chen, Tianqi and Guestrin, Carlos},
title = {{XGBoost}: A Scalable Tree Boosting System},
year = {2016},
isbn = {9781450342322},
booktitle = {Proceedings of the 22nd ACM SIGKDD International Conference on Knowledge Discovery and Data Mining},
pages = {785–794},
numpages = {10},
}

@article{breiman2001random,
  title={Random forests},
  author={Breiman, Leo},
  journal={Machine learning},
  volume={45},
  number={1},
  pages={5--32},
  year={2001},
  publisher={Springer}
}

@article{10.1145/3626315,
author = {Huawei, Huang and Qinnan, Zhang and Taotao, Li and Qinglin, Yang and Zhaokang, Yin and Junhao, Wu and Xiong, Zehui and Jianming, Zhu and Wu, Jiajing and Zheng, Zibin},
title = {Economic Systems in the Metaverse: Basics, State of the Art, and Challenges},
year = {2023},
volume = {56},
number = {4},
issn = {0360-0300},
journal = {ACM Computing Surveys},
articleno = {99},
numpages = {33},
}

@article{MOTIE2024122156,
title = {Financial fraud detection using graph neural networks: A systematic review},
journal = {Expert Systems with Applications},
volume = {240},
pages = {122156},
year = {2024},
issn = {0957-4174},
author = {Soroor Motie and Bijan Raahemi}
}

@article{Cheng_2025,
   title={Graph neural networks for financial fraud detection: a review},
   volume={19},
   ISSN={2095-2236},
   number={9},
   journal={Frontiers of Computer Science},
   author={Cheng, Dawei and Zou, Yao and Xiang, Sheng and Jiang, Changjun},
   year={2025}}

@ARTICLE{10026513,
  author={Huang, Yan and Li, Yi Joy and Cai, Zhipeng},
  journal={Big Data Mining and Analytics}, 
  title={Security and Privacy in Metaverse: A Comprehensive Survey}, 
  year={2023},
  volume={6},
  number={2},
  pages={234-247}}

@ARTICLE{10550925,
  author={McAmis, Rachel and Durak, Betül and Chase, Melissa and Laine, Kim and Roesner, Franziska and Kohno, Tadayoshi},
  journal={IEEE Security \& Privacy}, 
  title={Handling Identity and Fraud in the Metaverse}, 
  year={2025},
  volume={23},
  number={1},
  pages={27-37}}

@article{KARAPATAKIS2025100118,
title = {Metaverse crimes in virtual (Un)reality: Fraud and sexual offences under English law},
journal = {Journal of Economic Criminology},
volume = {7},
pages = {100118},
year = {2025},
issn = {2949-7914},
author = {Andreas Karapatakis},
}

\end{document}